\documentclass[man, floatsintext]{apa7}
\usepackage{float} 
\usepackage[style=apa,sortcites=true]{biblatex}
\usepackage{setspace}
\usepackage{amsmath}

\title{Large Language Model-Informed Feature Discovery Improves Prediction and Interpretation of Credibility Perceptions of Visual Content}

\shorttitle{LLM-Informed Feature Discovery for Credibility Perceptions}

\makeatletter
\renewcommand{\maketitle}{
\begin{titlepage}
    \centering
    \vspace*{2cm}
    {\bfseries\Large \@title\par}
    \vspace{12pt}
    
    {\normalsize
    Yilang Peng\textsuperscript{1}, Sijia Qian\textsuperscript{2}, Yingdan Lu\textsuperscript{3}, Cuihua Shen\textsuperscript{2}\par}
    
    \vspace{12pt}
    
    {\footnotesize
    \textsuperscript{1}Department of Financial Planning, Housing and Consumer Economics, University of Georgia\\
    \textsuperscript{2}Department of Communication, University of California, Davis\\
    \textsuperscript{3}Department of Communication Studies, Northwestern University\par}
    
    \vfill
    
    {\footnotesize
    Corresponding author: Yilang Peng, yilang.peng@uga.edu}
\end{titlepage}

\thispagestyle{plain}
\begin{center}
    \bfseries \@title
\end{center}
}

\makeatother

\begin{document}

\maketitle

\section{Abstract}
In today’s visually dominated social media landscape, predicting the perceived credibility of visual content and understanding what drives human judgment are crucial for countering misinformation. However, these tasks are challenging due to the diversity and richness of visual features. We introduce a Large Language Model (LLM)-informed feature discovery framework that leverages multimodal LLMs, such as GPT-4o, to evaluate content credibility and explain its reasoning. We extract and quantify interpretable features using targeted prompts and integrate them into machine learning models to improve credibility predictions. We tested this approach on 4,191 visual social media posts across eight topics in science, health, and politics, using credibility ratings from 5,355 crowdsourced workers. Our method outperformed zero-shot GPT-based predictions by 13\% in $R^2$, and revealed key features like information concreteness and image format. We discuss the implications for misinformation mitigation, visual credibility, and the role of LLMs in social science.

\noindent\textbf{Keywords:} Credibility, Large Language Model, GPT, Visual Media, Feature Discovery


\newpage

\setlength{\parindent}{0.5in}
\setlength{\parskip}{0pt}
\doublespacing

\clubpenalty=10000  
\widowpenalty=10000 

\section{Introduction}

Credibility perceptions are essential to citizens’ informed decision-making and are increasingly shaped by visual content, such as images and videos, in today’s media landscape \parencite{newman2024misinformed,peng2023agenda,guilbeault2024online}. Compared to textual information, visual modalities often improve the perceived credibility of a message, drawing on the “seeing is believing” heuristic and the richness of visual content \parencite{sundar2021seeing,ou2024factors}. Additionally, visuals tend to be more effective than text in capturing attention, increasing engagement, and improving memory retention \parencite{houts2006role, casas2019images, lu2024mobilizing}. Yet, decontextualized or manipulated visuals can support misleading claims, distort reality, and erode trust in the media \parencite{dobber2021microtargeted,vaccari2020deepfakes,shen2019fake,qian2023fighting}. 

Predicting credibility perceptions of social media visual content and understanding what makes visuals appear credible are critical for countering misinformation and prioritizing high-risk posts. However, these tasks remain challenging due to the diversity and complexity of the content features involved. These features span multiple levels, from basic visual elements like brightness and color to more subjective and difficult-to-quantify qualities such as aesthetic appeal, objectivity, and comprehensibility \parencite{peng2023agenda,sundar2008main}. This task is further complicated by the variety of visual formats, such as graphics and memes, as well as the multimodal nature of credibility perceptions, since visuals on social media are often accompanied by captions. In turn, current research on the credibility perceptions of social media visuals has largely focused on sources, heuristics, modalities, and individual differences rather than examining specific visual features \parencite{peng2023agenda,shen2019fake}.

Advancements in deep learning and computer vision (CV) have enabled social scientists to examine subjective concepts embedded in visual data, such as media bias, sentiment, and aesthetic appeal \parencite{joo2022image, peng2018same,talebi2018nima,iigaya2021aesthetic}. Scholars can adopt a top-down approach by selecting and designing features to predict these concepts, such as detecting faces and objects to study politicians’ portrayals or analyzing color and composition for aesthetic evaluation \parencite{joo2014visual,peng2018same,joo2022image,iigaya2021aesthetic}. However, this approach relies heavily on domain expertise and may also lead to incomplete feature selection. Alternatively, researchers can apply deep learning techniques to predict image perceptions and use methods like activation mapping to locate image regions that contribute to these predictions \parencite{talebi2018nima, joo2022image}. While this approach effectively identifies relevant objects or details, it can be challenging to connect these local visual patterns to broader content attributes, potentially limiting interpretability.

With the advent of visual processing capabilities, multimodal large language models (LLMs)—also referred to as vision-language models (VLMs)—offer a promising approach for understanding and reasoning about visual content. In social science research, LLMs have been rapidly adopted for textual analysis—often matching or even surpassing human annotators in identifying social, psychological, and political concepts, even in zero-shot settings, where they make predictions without task-specific training \parencite{gilardi2023chatgpt,rathje2024gpt}. However, their application to visual data remains relatively underexplored. Early research suggests that while LLMs like GPT achieve human-level performance in some visual tasks, it struggles with other domains such as causal reasoning \parencite{schulze2025visual,alexander2024can}. Research is needed to evaluate LLMs’ ability to assess visual media attributes, such as perceived credibility, and to develop an interpretable framework for understanding the content features that drive these evaluations.

\begin{figure}[tbhp]
    \centering
    \includegraphics[width=1\textwidth]{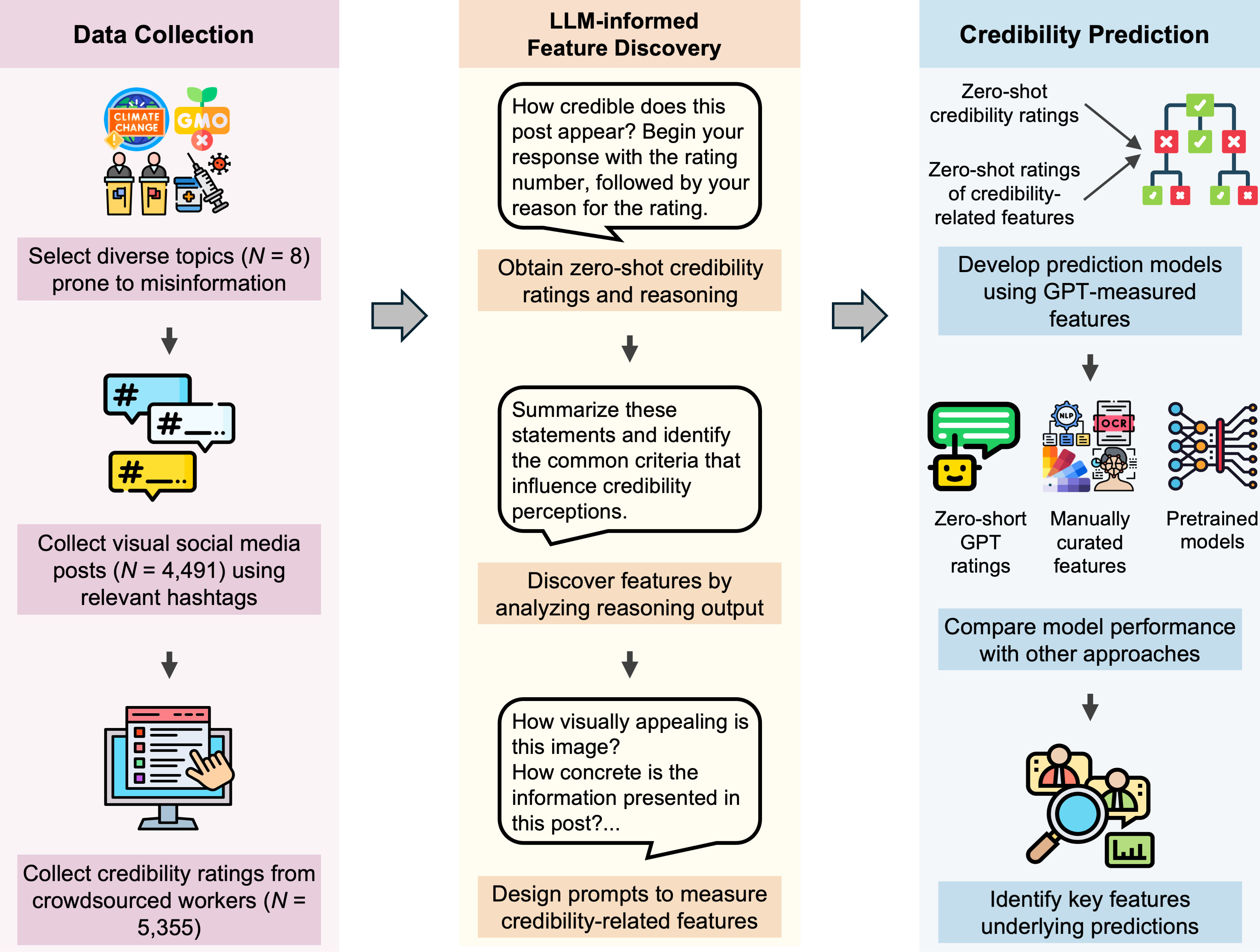}
    \caption{Workflow of the study. In our proposed LLM-informed feature discovery framework, we leveraged GPT-4o’s reasoning capabilities to identify potential features underlying credibility perceptions, designed prompts to explicitly measure these features, and incorporated them to predict human credibility judgments. This approach improves both the accuracy and interpretability of credibility predictions.}
    \label{fig:workflow}
\end{figure}

Addressing these gaps, we propose a LLM-informed feature discovery framework—a multistage, human-in-the-loop approach for analyzing subjective content evaluations, such as perceived visual credibility. We demonstrate that: (1) multimodal LLMs like GPT-4o can effectively mimic human credibility perceptions, and (2) when used for feature discovery and measurement, they improve the prediction and interpretation of visual content credibility beyond zero-shot methods. Our work provides theoretical insights on what contributes to the perceived credibility of visual media. Methodologically, it demonstrates how to leverage LLMs’ ability to measure and interpret subjective concepts in social science beyond zero-shot approaches.

\section*{LLM-informed Feature Discovery Framework}
Our proposed framework (Fig.~\ref{fig:workflow}) leverages the unique strengths of multimodal large language models to uncover and operationalize features associated with the perceived credibility of social media content. Specifically, it draws on three key capabilities of LLMs: (1) their ability to process both visual and textual information, (2) their capacity to perform zero-shot evaluations of content features, and (3) their reasoning abilities that provide interpretable explanations for their assessments. To demonstrate and evaluate the framework, we first collected a comprehensive dataset of 4,191 image-based social media posts from Twitter/X, along with credibility perceptions from crowdsourced workers. These posts cover eight issues in diverse contexts: climate change, GMO food, COVID-19, vaccines, general health, 2020 U.S. election, the Russo-Ukrainian War, and the Israeli–Palestinian conflict.

A key innovation of our approach is the use of GPT-4o’s reasoning output to discover potential credibility-related features. We first used GPT-4o to assess the credibility perceptions of social media posts in a zero-shot manner. We instructed it to provide a numeric score, explain its reasoning, and consider both the caption and the image in its evaluation. Then, by systematically analyzing and summarizing GPT-4o’s explanations, we extracted common themes and identified key criteria that underpinned its credibility judgments, such as sensational content, caption formality, and aesthetic quality. This LLM-informed feature discovery approach offers a structured and comprehensive way to identify credibility-related attributes that might otherwise be overlooked by traditional approaches.

Building on these insights, the next phase of our framework involves explicit feature quantification using LLMs. We crafted targeted prompts to instruct GPT-4o to quantify these credibility-related attributes. These quantified features were then used as inputs for predictive modeling of human-perceived credibility. We applied machine learning algorithms—such as Random Forest—to evaluate model performance and identify the most influential features driving credibility judgments. This integrated pipeline not only enhances the accuracy of credibility predictions but also provides interpretable insights into the factors shaping human perceptions.

\begin{figure}[tbhp]
\centering
\includegraphics[width=\linewidth]{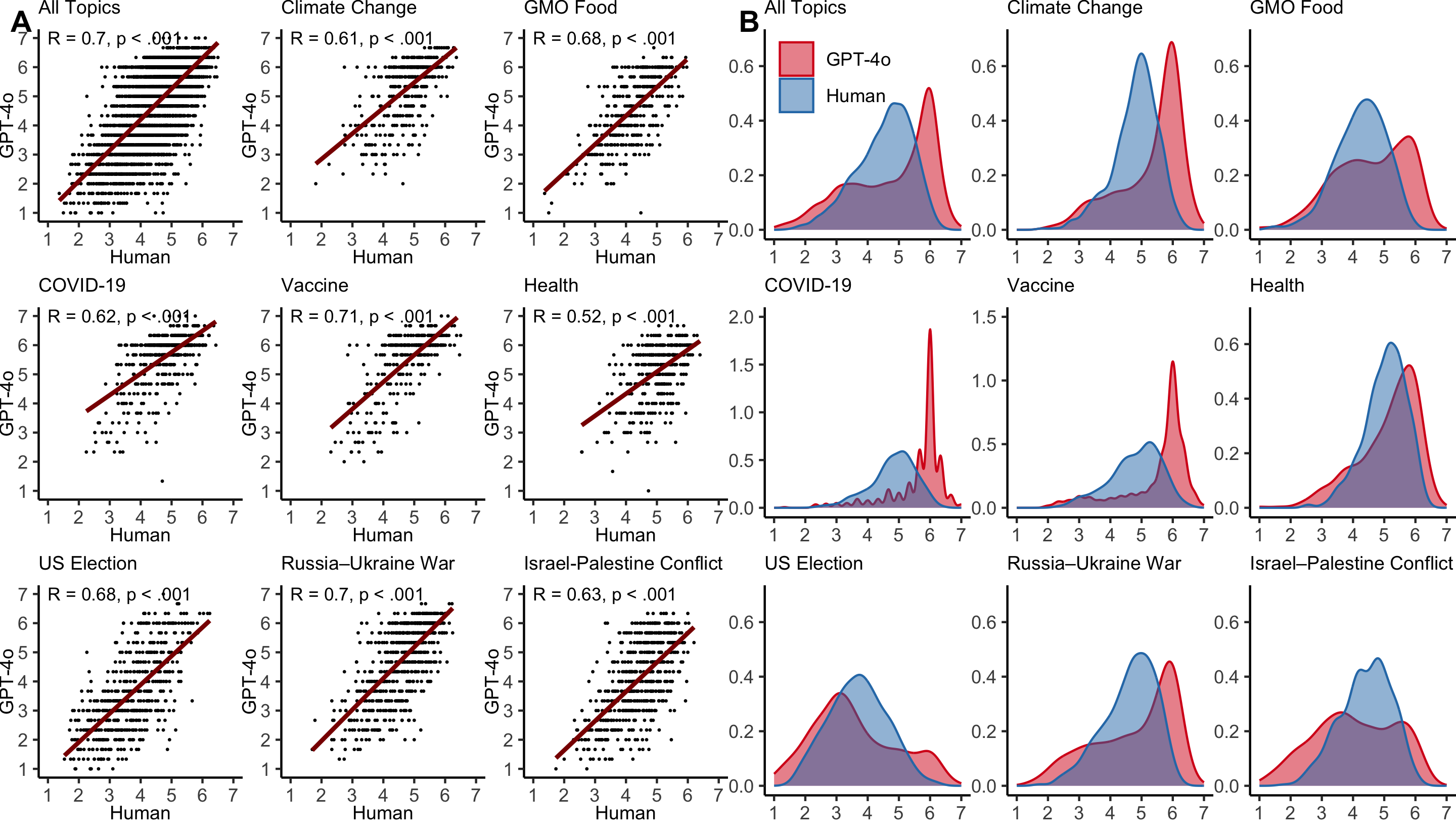}
\caption{(\textit{A}) Scatterplots illustrating the correlation between zero-shot GPT-4o–predicted credibility scores and human-perceived credibility scores (averaged at the post level). Results are presented both across all topics and separately for each individual topic. Scores range from 1 to 7, with higher values indicating greater perceived credibility and 4 representing a neutral midpoint.
(\textit{B}) Density plot showing the distribution of credibility scores from both GPT-4o predictions and human ratings.}
\label{fig:corr}
\end{figure}

\section*{Results}
\subsection*{GPT-4o’s Zero-Shot Credibility Ratings Largely Aligned with Human Judgments}
We began by evaluating GPT-4o’s zero-shot predictions of perceived credibility as a baseline. Overall, GPT-4o’s ratings demonstrated a relatively strong correlation with the average human credibility ratings at the post level ($r = 0.7$, $p < .001$, Fig.~\ref{fig:corr}\textit{A}). This association remained strong across specific topics, including GMO food ($r = 0.68$), vaccines ($r = 0.71$), the 2020 U.S. presidential election ($r = 0.68$), and the Russo-Ukrainian War ($r = 0.7$). However, in areas such as health ($r = 0.52$, all \textit{p}s $< .001$), the correlation was lower. Thus, GPT-4o predictions generally align well with human perceptions across a variety of topics, though the strength of this alignment varies by issue.

Despite the strong correlation, some discrepancies existed between the distributions of GPT-4o and human scores (Fig.~\ref{fig:corr}\textit{B}). Overall, GPT-4o tended to give higher scores (\textit{M} = 4.76, \textit{SD} = 1.35) than human ratings (\textit{M} = 4.53, \textit{SD} = 0.9; paired \(t(4,190) = 15.7, p < .001\)). Both distributions were negatively skewed, but human ratings showed a flatter distribution (skew = -0.66, kurtosis = -0.67) compared to GPT-4o ratings (skew = -0.59, kurtosis = -0.08). However, this trend varied by issues and was particularly pronounced for COVID-19 and vaccines.


\begin{figure}[tbhp]
   \centering
\includegraphics[width=1\textwidth]{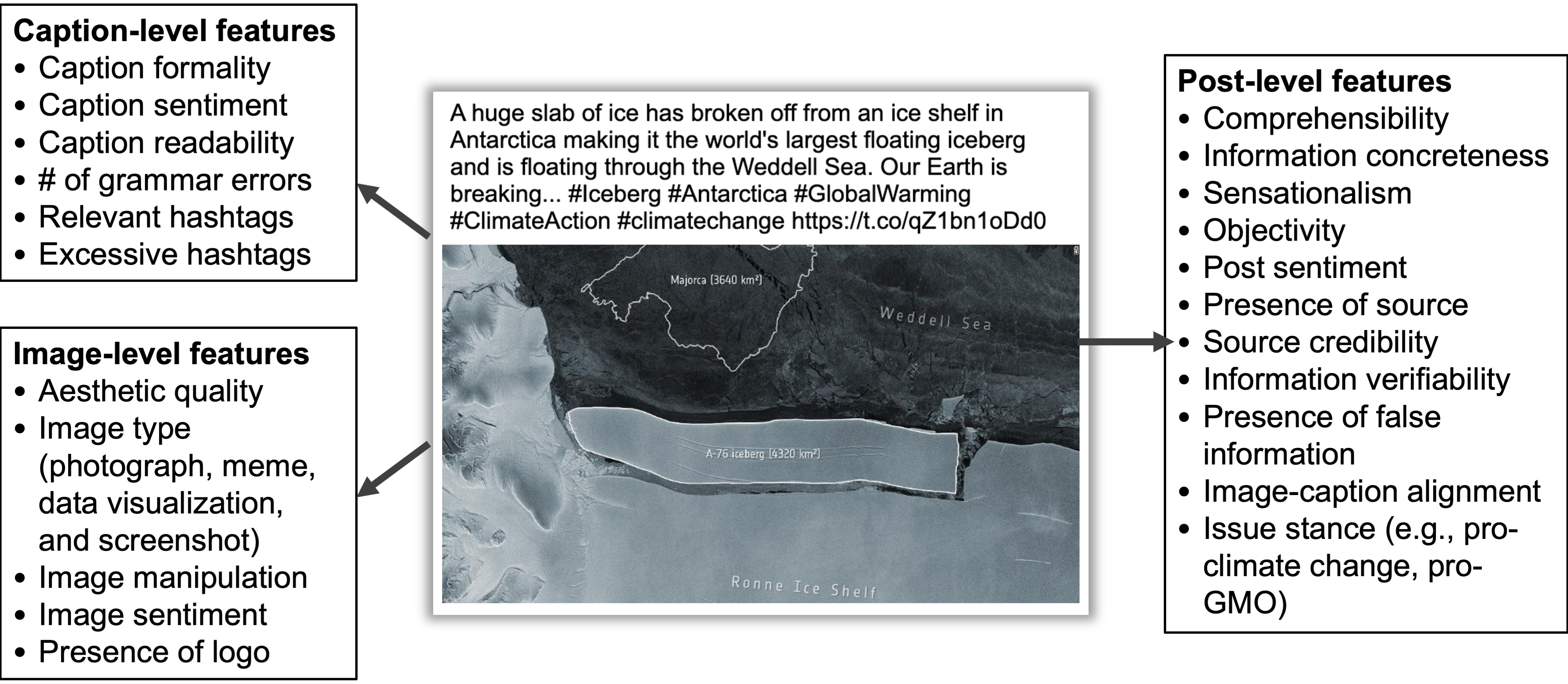}
   \caption{Overview of GPT-measured features used for analyzing social media posts, categorized into caption-level, image-level, and post-level features. The central image illustrates an example of a survey stimulus used for crowdsourcing ratings, where participants viewed a caption alongside the corresponding image below.}
    \label{fig:featurelist}
\end{figure}

\subsection*{GPT-4o’s Reasoning Revealed Features Underlying Credibility Perceptions}

By analyzing GPT-4o’s reasoning, we identified key content features influencing its credibility ratings, such as formal tone, concreteness, sensational content, and objectivity. Some of these factors aligned with prior research on credibility perceptions across domains \parencite{sundar2008main,peng2023agenda,pelau2023breaking,schwarz2021metacognitive,robins2008aesthetics}. For example, GPT-4o rated posts with clear, comprehensible language and visuals as more credible, consistent with research on processing fluency, which links ease of understanding to perceived credibility \parencite{schwarz2021metacognitive,ou2024factors}. Likewise, concrete information enhanced credibility, echoing findings that detail and specificity strengthen narrative credibility \parencite{nadel2024perceived}. GPT-4o also associated high aesthetic and professional image quality with greater credibility, in line with research showing that well-designed websites are perceived as more credible \parencite{robins2008aesthetics} and that online reviews featuring aesthetically appealing images are considered more helpful \parencite{pan2024aesthetic}. Also, the alignment between captions and images contributed to GPT-4o's credibility assessment, consistent with research showing that textual-visual similarity enhances the perceived helpfulness of reviews \parencite{ceylan2024words}. Still, GPT-4o’s reasoning suggested additional features that have received less attention in prior research, such as the use of relevant and excessive hashtags and information verifiability—whether a post provided sufficient context or background to enable and facilitate verification (for the definitions of all features, see \textit{SI Appendix} Table S3). 

We further categorized these features into three levels: caption-level, image-level, and post-level (Fig.~\ref{fig:featurelist}). Caption-level features included attributes directly related to captions, such as grammatical errors and readability. Image-level features described characteristics of the images, such as their format (e.g., memes), aesthetic quality, and image manipulation. Post-level features captured overall assessments of the post, such as comprehensibility and sensationalism. To capture these factors underlying credibility perceptions, we designed targeted prompts to instruct GPT-4o to explicitly measure these features (\textit{SI Appendix}, Appendix B).

\begin{figure}[tbhp]
   \centering
\includegraphics[width=1\textwidth]{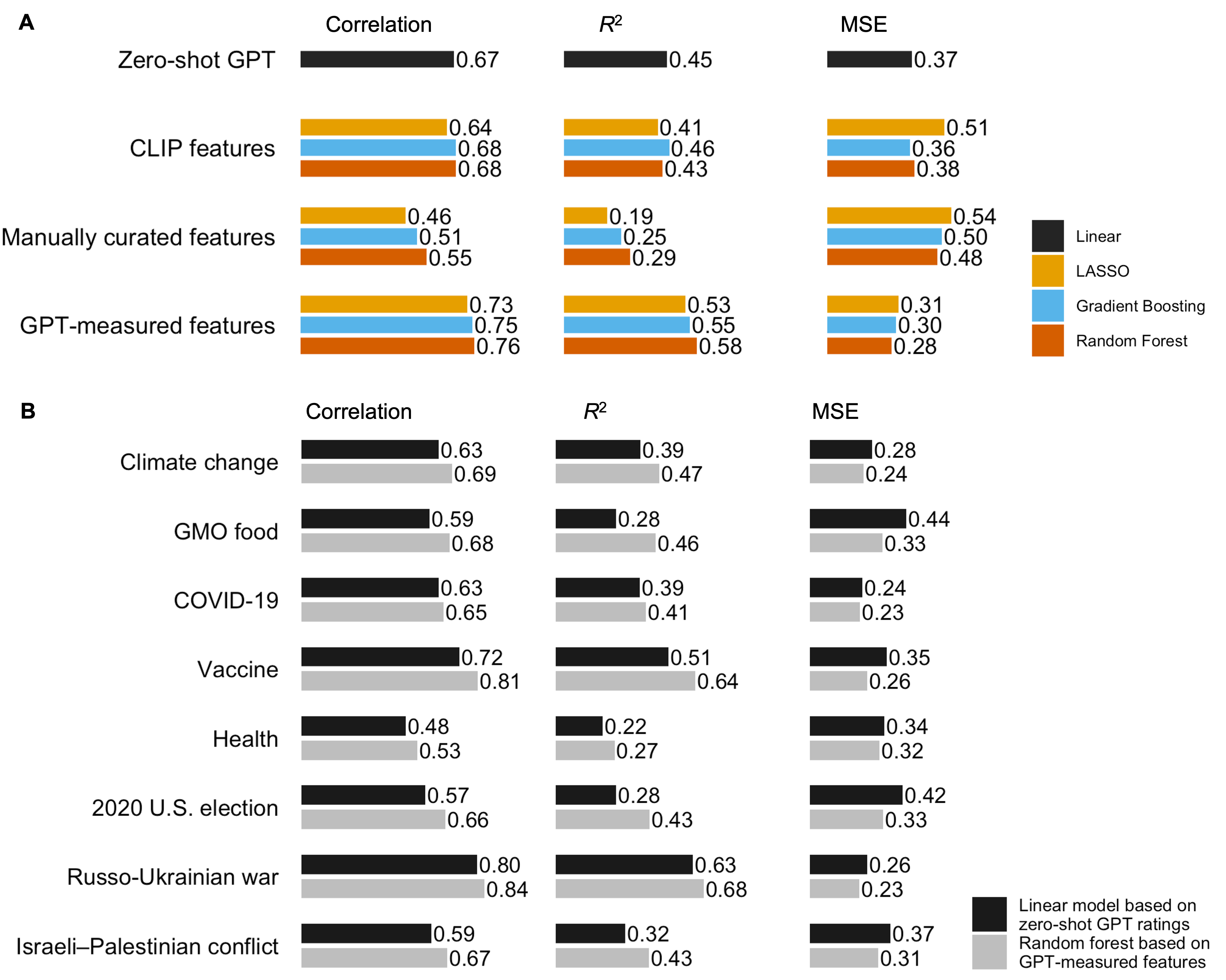}
  \caption{(\textit{A}) Comparison of machine learning models in predicting credibility perceptions for all issues combined. MSE = Mean Squared Error. All models were evaluated on a held-out test set (\textit{N} = 400) not used during training. The bottom two models additionally included binary indicators for the eight issues as features. (\textit{B}) Issue-specific comparison of a Random Forest model using GPT-measured features versus GPT-4o’s zero-shot credibility ratings. The linear model was trained separately within each issue, and evaluation was performed on that issue’s test set (\textit{N} = 50).
}
    \label{fig:compare}
\end{figure}

\subsection*{Incorporating GPT-measured Features Enhanced Credibility Predictions}

We compared our proposed approach—which integrated GPT-4o’s zero-shot credibility ratings along with its assessments of other credibility-related features to model human credibility perceptions—against three baseline models (Fig.~\ref{fig:compare}\textit{A}). The first baseline relied on a linear transformation of GPT-4o’s zero-shot credibility ratings. The second baseline used a set of computationally measured features, manually curated by researchers, from natural language processing (NLP) and CV methods, such as emojis, readability, color variations, visual complexity, and human faces. The third baseline relied on features from the Contrastive Language-Image Pre-training (CLIP) model, a pre-trained multimodal deep learning model \parencite{radford2021learning} (see \textit{Methods}). 

Among the three baseline models, the one trained on zero-shot GPT-4o ratings performed relatively well on the test dataset ($r = 0.67$, $R^2 = 0.45$, Mean Squared Error (MSE) = 0.37), demonstrating similar performance to the best-performing model trained on CLIP features ($r = 0.68$, $R^2 = 0.46$, MSE = 0.36). However, the model based on manually curated features performed worse, with the best-performing Random Forest model achieving $r = 0.55$, $R^2 = 0.29$, and MSE = 0.48.

Our proposed method outperformed all baselines. A Random Forest model using GPT-4o features achieved $r = 0.76$, $R^2 = 0.58$, and $\text{MSE} = 0.28$. Compared to the zero-shot GPT model, $R^2$ increased by 13 percentage points (a 28.9\% relative improvement), while $\text{MSE}$ decreased by 0.09 (a 24.3\% relative reduction).

We also investigated whether this improvement was consistent across all eight topics examined. To do so, we compared the best-performing model—the Random Forest model trained on GPT-4o-measured features—with the linear regression model based on GPT-4o’s zero-shot credibility ratings (see \textit{Method}). As shown in Fig.~\ref{fig:compare}\textit{B}, the improvement in predictive performance was consistent across all topics. The most substantial gains were observed for the GMO topic, where the $R^2$ increased by 18\% (from 0.28 to 0.46), followed by the 2020 U.S. election with a 15\% increase (from 0.28 to 0.43) and vaccines with a 12\% increase (from 0.51 to 0.64). The smallest improvements were found for COVID-19 (2.7\% increase, from 0.39 to 0.41) and the Russo-Ukrainian War (4.9\% increase, from 0.63 to 0.68). These results demonstrate that our LLM-informed feature discovery framework offers consistent and generalizable enhancements in predicting human-perceived credibility across diverse issue areas.

\begin{figure}[tbhp]
    \centering
    \includegraphics[width=\textwidth]{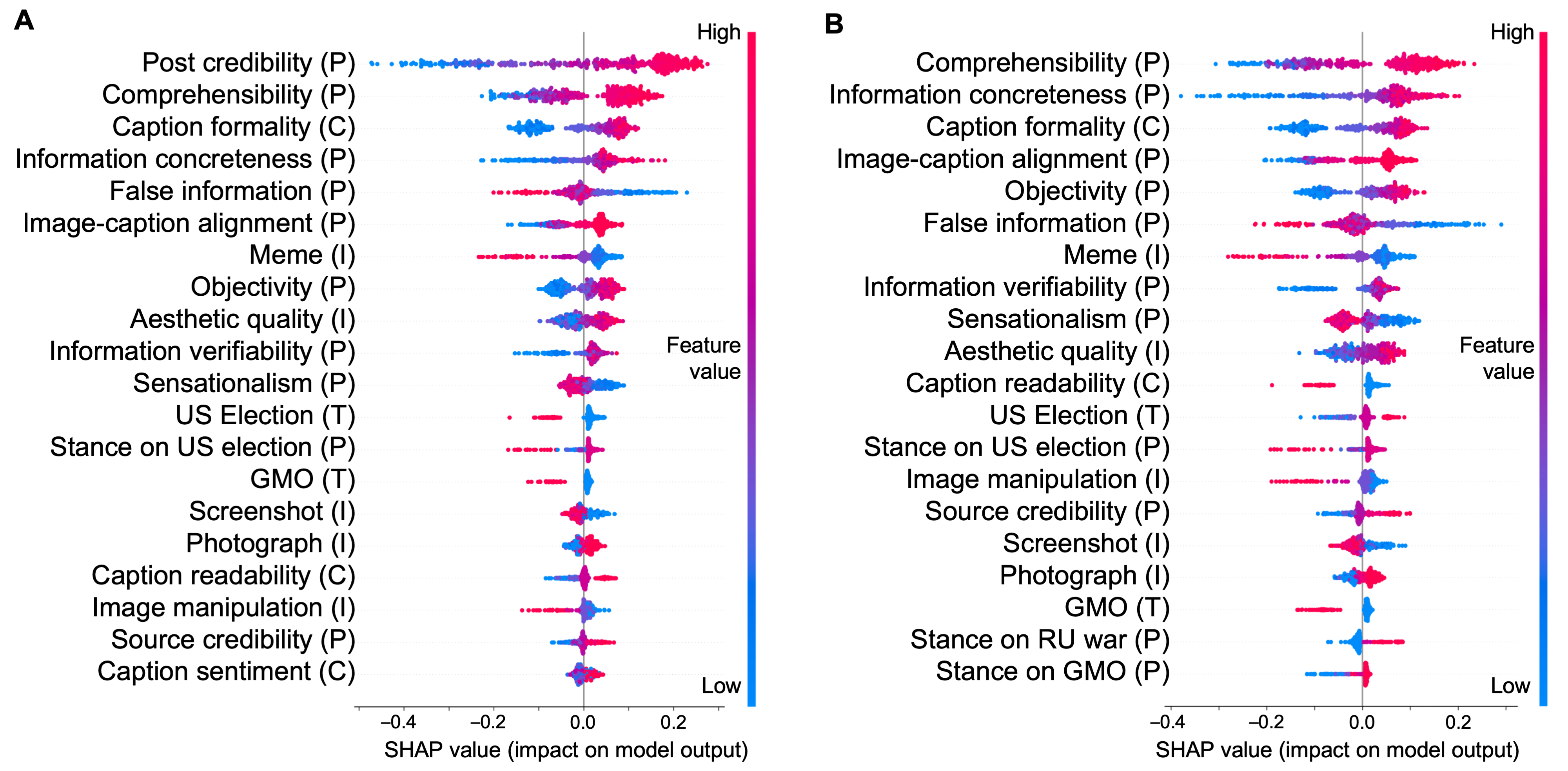}
    \caption{(\textit{A}) SHAP values for the top 20 features in a Random Forest model using GPT-measured features to predict human-perceived credibility. (\textit{B}) SHAP values for the top 20 features in a Random Forest model excluding GPT-4o’s zero-shot post credibility ratings. Features are grouped by type: post-level (P), caption-level (C), image-level (I), and topic indicators (T). Wider dot distributions indicate higher feature impact on the model's output. The color gradient represents feature values—red points on the right contribute positively to credibility perception, while blue points indicate a negative effect.}
    \label{fig:shap}
\end{figure}

\subsection*{Machine Learning Uncovered Key Features Predicting Human-Perceived Credibility}
Moreover, relying on zero-shot ratings or pre-trained models offered relatively limited insight into the factors that make a post appear credible to humans, highlighting a key limitation of these approaches. In our framework, we used SHAP (SHapley Additive exPlanations) values \parencite{lundberg2017unified} to analyze feature importance in predicting credibility perceptions. In our best-performing model that relied on GPT-measured features, GPT’s credibility rating emerged as the strongest predictor (Fig.~\ref{fig:shap}\textit{A}). This pattern confirmed that posts deemed credible by GPT-4o also tended to be rated as more credible by humans, indicating a strong alignment. However, this feature alone did not clarify what made a post credible and could overshadow the explanatory power of other credibility-related features. Therefore, we constructed an alternative model that excluded GPT-4o’s credibility rating (Fig.~\ref{fig:shap}\textit{B}).

Although this adjusted model showed a slight decrease in accuracy, it remained comparable overall, achieving $r = 0.75$, $R^2 = 0.55$, and $\text{MSE} = 0.30$ using a Random Forest approach. As revealed by SHAP values, post-level features—such as comprehensibility, information concreteness, image-caption alignment, objectivity, and verifiability—predicted higher perceived credibility, while the presence of false information negatively affected predictions. Among image-level features, higher aesthetic quality was associated with higher predictions, while manipulated images and memes reduced it. Captions that were formal and easy to read also predicted higher perceived credibility.

\section*{Discussion}
In summary, we demonstrated that our proposed LLM-informed feature discovery framework enhances the predictive accuracy of human-perceived credibility of visual content, while also offering interpretable insights into the most important features driving these predictions. Our findings carry several key implications for the use of LLMs in social science research. First, it underscores the potential of LLMs to approximate human credibility judgments of visual content. This aligns with previous findings that LLMs can effectively measure social science concepts in textual data, even in zero-shot settings \parencite{gilardi2023chatgpt,rathje2024gpt}. Moreover, our research advances this field by showing that recent multimodal models can also effectively annotate visual content. In \textit{SI Appendix} (Section 4 and Table S5), we provide further validation that GPT-4o can relatively accurately capture some of the credibility-related features for visual media, such as visual formats and aesthetic quality. With the rise of visually centric social media platforms and the growing prevalence of visual content \parencite{guilbeault2024online,newman2024misinformed,peng2024automated}, text-based analysis alone may not fully capture the contemporary media landscape. By using multimodal LLMs and incorporating visual data, social scientists can gain a comprehensive understanding of the information environment.

Furthermore, relying solely on LLMs’ zero-shot predictions may offer limited insight into how content features influence judgments. Our work demonstrates that LLMs’ reasoning capabilities can support systematic feature discovery: prompting GPT-4o to explain its decisions revealed meaningful features for machine learning pipelines. While prior research has proposed various credibility-related features, GPT-4o offers a relatively comprehensive and structured method for identifying features underlying credibility perceptions of visual social media posts. Because these features draw from diverse research areas and message types, this approach helps mitigate the risk of incomplete feature selection. In the \textit{SI Appendix} (Section 5 and Appendix C), we present a complementary analysis of qualitative interviews in which participants were asked to articulate what makes them perceive social media posts as more or less credible. The findings show that many features identified by GPT-4o were also mentioned by participants, who similarly relied on cues such as caption formality and sensational content when evaluating credibility. In summary, compared with traditional methods of feature discovery—such as relying on domain experts or conducting time-intensive qualitative interviews—our framework offers a scalable, efficient alternative. This underscores LLMs’ potential to enhance scalable, systematic feature discovery in content evaluation.

Our study offers insights into research on credibility and media effects. First, we present a comprehensive roadmap for human credibility evaluation at the post, image, and caption levels, guiding future investigations into content attributes and credibility perceptions. While some features, such as language formality \parencite{mosquera2012smile}, sensationalism \parencite{pelau2023breaking}, and concreteness \parencite{nadel2024perceived} have been explored in previous research, our roadmap highlights additional factors that merit further investigation. Moreover, this LLM-informed feature discovery framework should extend beyond credibility to other subjective evaluations. For instance, since concreteness plays a key role in shaping credibility perceptions, future research can adopt our approach to examine what makes information appear concrete versus vague and identify specific contributing features. Ultimately, this study suggests that LLMs can serve as a valuable tool for hypothesis generation in social science and provides a workflow for validating explanatory features \parencite{bail2024can}.

Our research provides practical implications for misinformation mitigation. First, this methodology helps identify visual posts that are most likely perceived as credible. This could assist platforms, fact-checkers, and public institutions in prioritizing detection and verification efforts. For example, screening for concrete information, formally written captions, and high-aesthetic images provides a low-cost yet effective method for identifying credible-looking misinformation that could have most impact. Moreover, our findings enhance media literacy efforts by identifying specific heuristics and biases that influence how people assess visual content. Media literacy education can incorporate these insights to teach users about cognitive biases and equip them with the skills to adopt a more analytical approach to online information and critically assess misleading yet credible-looking content \parencite{shen2019fake}. 

Still, it is crucial to acknowledge LLMs’ limitations. First, LLMs may amplify cognitive, cultural, and sociopolitical biases present in the data they learn from \parencite{zhou2024political,tao2024cultural}. This raises concerns about the fairness and generalizability of their outputs, especially when applied across diverse regions and contexts. Addressing these biases is critical to ensuring more reliable and equitable applications of LLMs in credibility assessment. In addition, LLMs are known for their tendency to hallucinate information \parencite{zhang2024risks}. While LLMs can generate sophisticated reasoning for their credibility judgments, the extent to which these factors actually influence credibility predictions remains unclear. This underscores the need to validate LLM-generated features against human perceptions to reveal the features that genuinely shape message assessments.

Future research should extend our study in several key directions. First, our study is limited by its reliance on GPT-4o, one of the most advanced LLMs with multimodal and reasoning capabilities available at the time. As LLMs continue to evolve, future research should evaluate the generalizability and replicability of our LLM-informed feature discovery framework for credibility assessment. Future research can also extend to a broader range of issues, contexts, and populations. For example, this study’s participants from Prolific may not fully represent the general population, as crowdworkers tend to be younger, more liberal, and more digitally literate \parencite{hargittai2020comparing}. Finally, while this study focuses on visual content features, individual differences are also crucial in credibility judgments \parencite{shen2019fake}. Future research can examine how user traits interact with content features and work toward developing personalized credibility prediction models.

\section{Methods}
\subsection*{Collection of Social Media Posts}
As shown in our study workflow (Fig.~\ref{fig:workflow}), we first identified topics that (1) were prone to misinformation, (2) included both polarizing and non-polarizing issues, and (3) spanned diverse domains. We included climate change, GMO food, COVID-19, vaccines, general health (such as nutrition and exercise), the 2020 U.S. presidential election, and the Russo-Ukrainian War. Following its escalation in October 2023, we added the Israeli-Palestinian conflict due to its high susceptibility to misinformation.

For each topic, we identified the top hashtags and manually coded them by stance to ensure balanced representation. Using relevant hashtags and the Twitter Academic API, we collected image posts from January 2018 to April 2022. For the Israeli-Palestinian conflict, we used a Python script to collect posts from October 23 to November 29, 2023. To ensure data quality, we manually filtered out irrelevant posts and removed duplicates and retweets. See \textit{SI Appendix}, Section 1 for a detailed description and Table S1 for post counts at each stage. 

\subsection*{Collection of Credibility Ratings}
Our final dataset for crowdsourced annotations included 4,191 posts covering eight topics: climate change (n = 474), GMO food (n = 428), COVID-19 (n = 595), vaccines (n = 376), general health (n = 417), the 2020 U.S. presidential election (n = 601), the Russo-Ukrainian War (n = 698), and the Israeli-Palestinian conflict (n = 602). We created mock social media posts using only the captions and images from original posts. To focus on content features, we removed all contextual information such as source, engagement metrics, and publication time to ensure they do not affect the judgment of content credibility (Fig.~\ref{fig:featurelist}). For posts containing multiple images, we generated a separate post for each individual image. 

We recruited U.S. participants via Prolific who had an approval rate above 95\% and had completed 200 tasks. Given the large number of annotation tasks, we published multiple batches, each including posts on a single topic, with data collection occurring at intervals of approximately two to three weeks between batches from December 2023 to June 2024. The final sample consisted of 5,355 participants, with 47.1\% identifying as male and an average age of 42.8 years (\textit{SD} = 13.5). Regarding political affiliation, 47.1\% identified as Democrats, 18.8\% as Republicans, and 26.9\% as Independents. Regarding education, 13.2\% of participants had a high school education or less, 27.9\% had some college education, 37.2\% held a bachelor’s degree, and 16.6\% had postgraduate education. Regarding race and ethnicity, 9.0\% of participants identified as Hispanic, 59.8\% identified as Non-Hispanic White, 13.5\% as Non-Hispanic Black, and 7.5\% as Non-Hispanic Asian. The participants’ political ideology was measured on a 7-point scale (1 = very liberal; 7 = very conservative; \textit{M} = 3.36, \textit{SD} = 1.77). 

In each survey, respondents first answered questions about their individual characteristics, issue relevance, and issue attitudes, followed by rating eight randomly selected social media posts. For each post, participants rated its perceived credibility by indicating to what extent the post was credible, accurate, and believable on a 7-point scale \parencite{appelman2016measuring}. On average, each post received 10.22 ratings (range = 8–13), with a total of 42,840 ratings collected. We averaged each post’s scores on perceived credibility, accuracy, and believability, then combined them into a single credibility rating for prediction tasks ($\alpha = .98$).

\subsection*{Feature Discovery with GPT-Reasoning}

We outline the steps in feature discovery in Fig.~\ref{fig:workflow}. First, we instructed GPT-4o to perform zero-shot estimations of the credibility of all samples and to provide reasoning for these estimations. In particular, we asked GPT-4o to evaluate the credibility of posts by rating how credible, believable, and accurate they appeared on a 7-point scale ($\alpha = .91$) and combined into one rating. For these assessments, we used GPT-4o version gpt-4o-2024-05-13 with the default temperature setting of 1.

We randomly selected 80 posts from the training set (10 posts for eight topic, see the following section on machine learning) along with GPT-4o’s reasoning. Using the following prompt, we asked GPT-4o to summarize key patterns in credibility perceptions: “Here is a list of statements about why certain social media posts are perceived as more or less credible. Summarize these statements and identify the common criteria that influence credibility perceptions.” This process was repeated for four additional batches of posts until no new themes or factors emerged from GPT-4o’s responses. This procedure revealed multiple key patterns that aid GPT-4o’s reasoning, such as professional tone and language, specificity and detail, emotional and sensational content, bias and objectivity (see \textit{SI Appendix}, Appendix A for an example).

We inspected and summarized these features into three levels: caption-level features, image-level features, and post-level features. To capture these features that underlie credibility perceptions, we designed targeted prompts to instruct GPT-4o to explicitly measure them. For objective features, such as visual format, we used a single prompt (e.g., assessing how likely an image is a meme). For more subjective attributes, we used multiple prompts. For example, for information concreteness, we asked GPT to evaluate how detailed, concrete, or specific the information in the post was. Regarding aesthetic quality of images, we asked GPT to assess to what extent the image was visually appealing, attractive, and of professional quality. We aggregated the resulting scores to represent each subjective feature, all of which demonstrated strong internal consistency (reliability scores > 0.7) The full set of prompts is available in the \textit{SI Appendix}, Appendix B.

For image-level features, we fed only images into GPT-4o; for caption-level features, only captions. For post-level features, both were provided as input. We included issues as features in the machine learning models, coded as eight dummy variables.

\subsection*{Machine Learning Models for Credibility Predictions}

To account for sample size, potential collinearity, and interactions among predictors, we employed three machine learning methods: LASSO regression, Random Forest, and Gradient Boosting decision trees. We divided our dataset into three subsets: a training set (\textit{N} = 3,391), a validation set (\textit{N} = 400), and a test set (\textit{N} = 400), with the validation and test sets each containing 50 randomly selected images from each of the eight topics.

Hyperparameter tuning was performed using the combined training and validation sets to identify optimal model configurations. For LASSO regression, this involved searching across a range of regularization strengths. Regarding Random Forest, we tuned parameters such as the number of trees, maximum tree depth, and the minimum number of samples required for node splitting and leaf formation. For Gradient Boosting, the tuning process included the number of boosting stages, maximum tree depth, learning rate, subsample ratio, and minimum samples per leaf.

Models were trained on the training set, and hyperparameters were selected based on the $R^2$ metric using the validation set. The final model was evaluated on the test set, which remained strictly reserved for performance assessment and was not used during training.

\subsection{Model Comparison and Evaluation}

We compared our approach to three baseline models to establish benchmarks and evaluated our proposed approach against these baselines. The first model was based on GPT-4o’s zero-shot ratings. While GPT-4o’s assessments correlated well with human ratings, discrepancies in rating ranges could lead to high prediction errors. Directly applying GPT-4o’s ratings resulted in poor performance, with a negative $R^2$ and a high MSE. We therefore applied a linear regression model to the combined training and validation set (as no tuning was needed).

For the second model, we manually curated a diverse set of computational features from CV and NLP methods. These features may underlie human evaluations of visual posts, such as aesthetic appeal and sentiment, which could shape credibility \parencite{peng2023agenda}. For images, we extracted color-related features (e.g., brightness, color distribution), composition features (e.g., aspect ratio, visual complexity), and quality metrics (e.g., sharpness). We also used deep learning models to assess aesthetic appeal and technical quality \parencite{talebi2018nima}, applied facial recognition to detect faces and emotions, and adopted OCR to quantify text presence. For captions, we analyzed word count, hashtags, mentions, emojis, URLs, readability, and sentiment (for the full list of features, see \textit{SI Appendix}, Table S4).

For the third comparison, we resorted to deep learning methods, which are widely used to analyze visual data. In social science research, where datasets are often small, pre-trained models help extract reusable features without the need for extensive training \parencite{joo2022image,peng2024automated,zhang2024image}. We used feature extraction with OpenAI CLIP model (ViT-B/32) \parencite{radford2021learning}, which was designed for both text and images, by extracting features from one of its final layers and feeding them into a machine learning classifier to predict credibility.

We evaluate the performance of different models on the test set (\textit{N} = 400) using three indicators: correlation, $R^2$, and MSE. This test set was also used to compute SHAP values.

Finally, to compare model performance across the eight topics, we trained eight separate linear regression models—one for each issue domain—to capture potential differences in credibility perceptions. Each model was trained on the combined training and validation sets specific to its issue domain and evaluated on the test set (\textit{N} = 50).

\subsection{Acknowledgement}
This work was supported by the U.S. National Science Foundation (CNS-2150716, CNS-2150723), OpenAI researcher access program, and Google Cloud research credits program.

\printbibliography
\end{document}